\documentclass[letterpaper, 10 pt, conference]{ieeeconf}  

\IEEEoverridecommandlockouts                              

\DeclareMathAlphabet{\mathcal}{OMS}{cmsy}{m}{n}

\usepackage{graphics} 
\usepackage{epsfig} 
\usepackage{mathptmx} 
\usepackage{times} 
\usepackage{amsmath} 

\usepackage{amssymb}  
\usepackage[T1]{fontenc}
\usepackage{caption}
\usepackage{subcaption}
\usepackage{tikz}
\usepackage{algorithm}
\usepackage{algpseudocode}

\usepackage{cite}
\usepackage{multirow}
\usepackage[normalem]{ulem}
\usepackage[percent]{overpic}
\usepackage{bbm}
\allowdisplaybreaks

\usepackage[markup=bfit, deletedmarkup=sout, authormarkup=superscript]{changes}
\definechangesauthor[name={ml}, color=blue] {ML}

\definechangesauthor[name={kb}, color=purple]{KB}

\definechangesauthor[name={ar}, color=red]{AR}

\definechangesauthor[name={ar}, color=brown]{YT}

\title{\LARGE \bf
Bilevel Optimization for Just-in-Time Robotic Kitting and \\Delivery via Adaptive Task Segmentation and Scheduling
}

\author{Yi-Shiuan Tung, Kayleigh Bishop, Bradley Hayes, and Alessandro Roncone
\thanks{*This work was supported by the Army Research Laboratory under grant W911NF-20-2-0083, W911NF-21-2-0290, and W911NF-21-2-0123, and by the Office of Naval Research under grant N00014-22-1-2482.}
\thanks{The authors are affiliated with the 
        Department of Computer Science,
        University of Colorado Boulder, 1111 Engineering Drive, Boulder, CO 80309, U.S.A.
        {\tt\small \{firstname\}.\{lastname\}@colorado.edu}}%
}

\begin{document}
\renewcommand{\arraystretch}{1.5}

\maketitle
\thispagestyle{empty}
\pagestyle{empty}

\begin{abstract}
Kitting refers to the task of preparing and grouping necessary parts and tools (or ``kits'') for assembly in a manufacturing environment. Automating this process simplifies the assembly task for human workers and improves efficiency. Existing automated kitting systems adhere to scripted instructions and predefined heuristics. However, given variability in the availability of parts and logistic delays, the inflexibility of existing systems can limit the overall efficiency of an assembly line. In this paper, we propose a bilevel optimization framework to enable a robot to perform task segmentation-based part selection, kit arrangement, and delivery scheduling to provide custom-tailored kits \textsl{just in time}---i.e., right when they are needed. We evaluate the proposed approach both through a human subjects study (n=18) involving the construction of a flat-pack furniture table and shop-flow simulation based on the data from the study. Our results show that the just-in-time kitting system is objectively more efficient, resilient to upstream shop flow delays, and subjectively more preferable as compared to baseline approaches of using kits defined by rigid task segmentation boundaries defined by the task graph itself or a single kit that includes all parts necessary to assemble a single unit. 
\end{abstract}

\section{INTRODUCTION}




In the conventional or ``single model'' assembly line, only one product type is manufactured at a time. While this traditional manufacturing approach simplifies the responsibilities of line workers, it limits flexibility: making one product at a time means costly changeovers when switching product types, making it difficult to respond fluidly to changing dynamics of customer demand, upstream part shortages, and customizations. For this reason, manufacturers may employ \textsl{``mixed model'' assembly}, in which multiple products --- with different parts and assembly steps --- are assembled on the same line. This approach smooths upstream part demand and can lead to much greater overall productivity in the assembly line \cite{zhu2008}.

The trade-off of mixed model assembly is that it requires assembly line workers to take on a greater range of tasks. Rather than constantly performing a rote series of assembly steps, workers now need to keep track of the various parts and steps for each product on the line. One approach to streamlining this step at the worker level is \textsl{kitting}. In manufacturing systems, kitting refers to the process of collecting components in a "kit" or container before feeding them to workstations where intermediate or end products are built. Kitting possesses a number of advantages: it 1) reduces storage space requirements at the workstations, 2) enables efficient product changeover (since common components are stored at a central location), and 3) increases worker productivity by reducing time spent gathering parts \cite{bozer1992kitting, fansuri2017challenges}.

\begin{figure}
\centering
\includegraphics[width=0.85\linewidth]{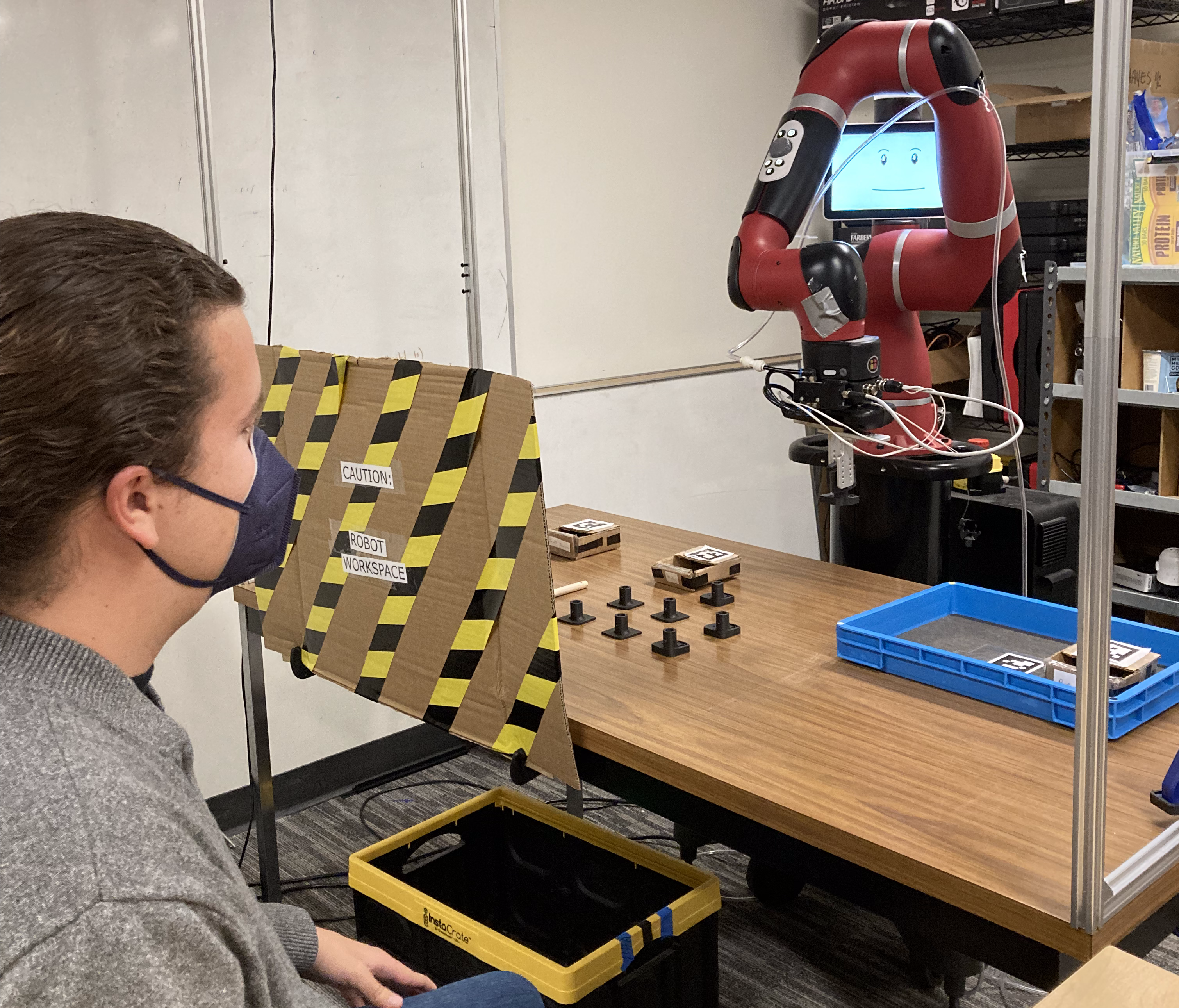}
\caption{This work presents an optimization approach to automate just-in-time robot part delivery for collaborative assembly. Here, a robot gathers parts required for an assembly task and places them on a kitting tray. The human then uses this "kit" to complete the next set of tasks for continuing to assemble the product.}
\label{fig: experiment setup}
\end{figure}
Kitting is an excellent opportunity space for leveraging collaborative robotics to improve the efficiency of workers. While automation has seen great success in many areas of manufacturing, assembly tasks often remain too complex to automate completely: robots lack the fine-motor capabilities required for dexterous manipulation of small parts and tools. Kitting, however, is well within the mechanical capabilities of modern robots, and its automation would allow a robot to prepare kits while more dexterous agents assemble the product. This parallelism could greatly decrease the time and effort required to complete each product assembly.

In this work we propose and evaluate a dynamic robotic kitting planner that segments and schedules assembly tasks to minimize idle time and reduce makespan. The primary contributions of this work are: a) a novel optimization algorithm for just-in-time robotic kitting, which incorporates a model of the assembly goal, estimates of human and robot task times, and estimated usability of the generated kit layout; b) a demonstration of the optimization and its performance outcomes on an experimental human-robot collaboration scenario; and c) an analysis of the optimization's robustness to logistic delays in simulation.

We formulate just in time kitting and delivery as a \textsl{bilevel} optimization, where the ``upper-level'' problem of task scheduling and segmentation is optimized in concert with the ``lower-level'' problem of part kitting. By performing this nested optimization, the robot adapts to its kitting conditions (e.g. kit arrangement and part availability)  while syncing its workflow to that of the human.

To evaluate our kitting approach's ability to increase efficiency and improve users' experience in a human-robot collaborative assembly, we implement our system on a Sawyer robot (Figure \ref{fig: experiment setup}) and conduct a within-subjects user study of collaborative furniture assembly. To evaluate the system's performance in scenarios with logistic delays, we use data from the user study to simulate the assembly process under varying conditions of part availability and throughput to explore our system's robustness.
%


\section{Background \& Related Work}\label{section: background}


One of the disadvantages of kitting is that the preparation of kits consumes worker time and effort that don't directly contribute to the assembly task. Kitting in mixed-model assembly lines also has higher rates of error \cite{jeanson2018tell} and assembly workers report higher cognitive loads \cite{joundi2019understanding,shaikh2012effects}.
%
Due to the physical and cognitive demands of manual kit preparation, robotic kitting has been explored to increase efficiency and flexibility. 
Caputo et al. \cite{caputo2021model} developed an economic model to show that automated systems are more cost effective and efficient, except in low production settings. 
Boudella et al. \cite{boudella2018kitting} developed a delivery-time-sensitive model to optimally distribute kitting tasks between a human and robot in a mixed-model assembly line, and saw significant efficiency improvements over baselines where the human worker took on the majority of the tasks. A mixed-integer linear programming approach to online kit delivery scheduling is presented in \cite{maderna2020online}, in which the robot takes over all kitting tasks and results in large productivity improvements over a manual kitting baseline.

Our work differs from the above examples in that it presents a fully-automated kitting system that achieves Just in Time (JIT) part feeding, a lean manufacturing principle designed to eliminate waste in production \cite{fansuri2017challenges}. The goal of JIT manufacturing is to feed the required quantity of parts to the workstations when they are needed, rather than before or after. Traditionally, JIT kitting is accomplished by determining the required kits at each workstation and designing the assembly layouts and kit sizes accordingly \cite{townsend2012line}. Other production systems use material handling techniques like kitting trolleys \cite{nikam2018design} or automated delivery vehicles \cite{zhou2021static} to achieve JIT. 

Several optimization approaches have been proposed to enable timely human robot or robot-robot interactions. An auction algorithm for solving task allocation with temporal constraints is presented in \cite{multi-robot-auction}, and a non-linear program for flexible scheduling of robot tasks is presented in \cite{wilcox2013optimization}. In addition to optimizing for efficiency, temporal considerations have been shown to enable fluent human robot teaming in \cite{time-aware-multiagent}. Our work differs from the above by using bilevel optimization to minimize idle times and also optimize for part selection and kit arrangement.


In this work, we use task segmentation and scheduling of kitting tasks to achieve JIT delivery. By incorporating an awareness of human and robot task times, as well as part availability into kit layout and design, our system is more capable of adapting its kitting behavior to meet the changing demands of mixed model assembly lines.

\section{Problem Formulation}\label{section: problem formulation}

Consider a two-station, human-robot collaborative task where a human assembler works at an assembly workstation and a robot prepares kits at a preparation workstation. The robot receives part deliveries and produces kits, while the human receives kits and produces product. 
Let:
$$\mathcal{A} = \{a_1, a_2,\ldots a_n\}$$
be the set of $n$ assembly tasks that need to be performed to complete the assembly. Each task $a_i$ is composed of the robot-dependent kitting task $a_i^{R}$, in which the robot gathers the necessary parts, and the human-dependent assembly task $a_i^{H}$. The robot has to complete preparation task $a_i^{R}$ (i.e., kitting and delivering components) before the human can begin $a_i^{H}$ (assembling). In order to schedule its actions, the robot's goal is to compute a \textsl{task ordering} over $\mathcal{A}$ for itself and the human to follow in order to finish the assembly as quickly as possible.

\begin{figure}
\begin{subfigure}[b]{0.48\linewidth}
\begin{tikzpicture}[sibling distance=1.8cm, every node/.style = {shape=rectangle, draw, align=center}, edge from parent/.style={draw,-latex}]
  \node [label=right:$a^H_1$] (t^H) {Insert screws\\to top}
    child { node[label=below:$a^H_2$] {Insert\\front legs} }
    child { node[label=below:$a^H_3$] {Insert\\back legs} };
\end{tikzpicture}
\caption{Human task}
\end{subfigure}
\begin{subfigure}[b]{0.48\linewidth}
\begin{tikzpicture}[sibling distance=1.8cm, every node/.style = {shape=rectangle, draw, align=center}, edge from parent/.style={draw,-latex}]
  \node[label=right:$a^R_1$] {Kit screws\\and top}
    child { node[label=below:$a^R_2$] {Kit\\front legs} }
    child { node[label=below:$a^R_3$] {Kit\\back legs} };
\end{tikzpicture}
\caption{Robot task}
\end{subfigure}
\caption{A task graph for assembly of a stool. The robot retrieves the parts required for the human to complete the assembly. The arrows indicate precedence constraints. The robot tasks are represented by $T^R = \{a^R_1, a^R_2, a^R_3\}$ and the human tasks are represented by $T^H = \{a^H_1, a^H_2, a^H_3\}$.}
\label{fig:example task graph}
\end{figure}
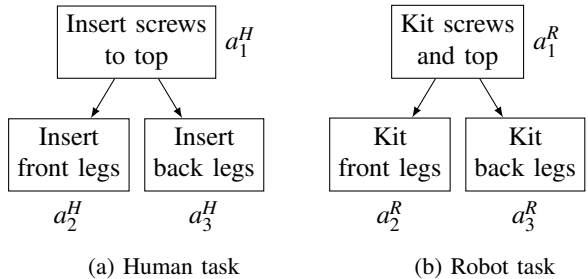

The tasks in $\mathcal{A}$ may not need to be completed in a strict order, but some tasks may have ordering constraints --- for example, one may need to complete ``insert screws to top'' before one can perform ``insert front leg" to build a table. Let us say that if task $a_i$ must be completed before $a_j$, then $a_i$ has a \textsl{precedence constraint} with $a_j$. Any task ordering the robot computes must satisfy these precedence constraints. See Figure \ref{fig:example task graph} for an example task precedence graph, and the associated part retrievals the robot needs to perform. Our proposed approach is generalizable to tasks with  precedence constraints as described above.

To create a \textsl{schedule} to complete each task given a task ordering, the robot \textsl{segments} tasks into groups. This means the robot can prepare a kit one segment at a time, rather than having to make one kit for each of the human's tasks. This also introduces the additional constraint that a segment can only be valid if all the associated parts can fit in the kit. Let us represent a task segment at time step $t$ as $T_t \subseteq \mathcal{A}$.

Once the tasks have been segmented into groups, the robot generates a schedule by determining when to start each segment. For this, it uses its prior knowledge for how long a given segment will take, both for itself and the human; let us represent this as $duration^H(T_t)$ and $duration^R(T_t)$ respectively. For the task in Figure \ref{fig:example task graph}, one possible segmentation is $T = \{T_1, T_2\} = \{\{a_1, a_2\}, \{a_3\}\}$ and a possible schedule is $S = \{0 s, 35 s\}$.
Lastly, the robot must arrange all of the parts for the current segment $T_t$ into the kit --- that is, all of the parts required for each task $a \in T_t$.

\section{Proposed Approach}\label{section: proposed approach}

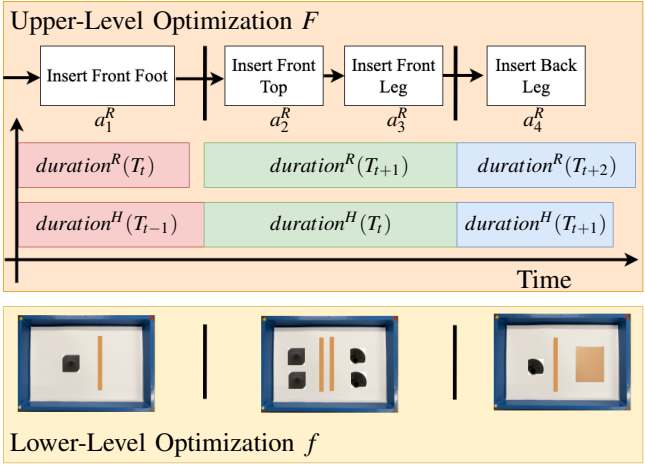
\begin{figure}
    \centering
    \begin{overpic}[width=\linewidth]{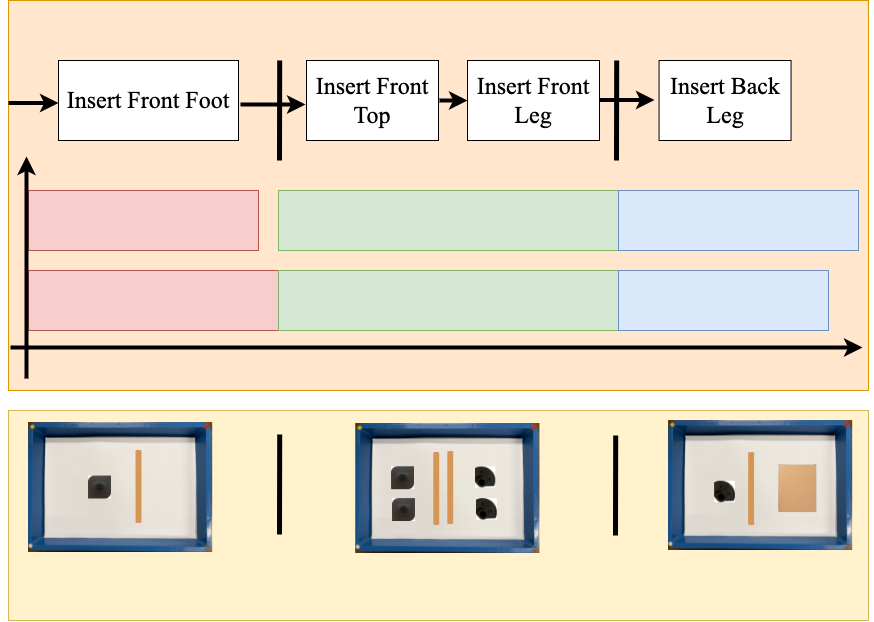}
        \put (6, 45) {\footnotesize $duration^R(T_{t})$}
        \put (6, 36) {\footnotesize $duration^H(T_{t-1})$}
        \put (42, 45) {\footnotesize $duration^R(T_{t+1})$}
        \put (42, 36) {\footnotesize $duration^H(T_{t})$}
        \put (74, 45) {\footnotesize $duration^R(T_{t+2})$}
        \put (72, 36) {\footnotesize $duration^H(T_{t+1})$}
        \put (15, 52) {\footnotesize $a^R_1$}
        \put (42, 52) {\footnotesize $a^R_2$}
        \put (60, 52) {\footnotesize $a^R_3$}
        \put (81, 52) {\footnotesize $a^R_4$}
        \put (80, 27) {Time}
        \put (2, 67) {Upper-Level Optimization $F$}
        \put (2, 2) {Lower-Level Optimization $f$}
    \end{overpic}
    \caption{Overview of the bilevel optimization algorithm for just in time kitting. The upper level optimization segments and schedules tasks, and the lower level optimization arranges parts on the tray. By minimizing the differences between $duration^R(T_{t+1})$ and $duration^H(T_{t})$ (shown in green) and the differences between $duration^R(T_{t+2})$ and $duration^H(T_{t+1})$ (shown in blue) as well as other terms in the objective function $F$ (Eq. \ref{eqn: precedence constraints} - \ref{term: lower-level}), the robot determined the task segment $T_{t+1}$ which corresponds to the set of tasks \textsl{Insert Front Top} ($a^R_2$) and \textsl{Insert Front Leg} ($a^R_3$). The task segment $T_{t+1}$ also depended on the kit fitness function $f$ (Eq. \ref{eqn: lower objective}) which evaluates the kit layout given the parts required $P$ for tasks in $T_{t+1}$. The robot's role is to prepare parts required for $a^R_2$ and $a^R_3$ in a kit according to the layout determined by $f$ at timestep $t+1$.}
    \label{fig:bilevel_optimization}
\end{figure}

We formulate our solution as a bilevel optimization, a hierarchical constrained optimization where one problem is embedded in another (i.e., Figure \ref{fig:bilevel_optimization}). The general formulation of a bilevel optimization problem is:
\begin{gather}
    \min_{x \in X, y \in Y} F(x, y) \quad s.t. \quad G(x, y) \leq 0\ \ , \label{bilevel:1} \\
    y \in \arg \min_{z \in Y}\{f(x, z) \quad s.t. \quad g(x, z) \leq 0 \}\ \label{bilevel:2}
\end{gather}

where $F$ is the upper-level objective function, $G$ represents the upper-level constraints, $f$ is the lower-level objective function, and $g$ represents the lower-level constraints \cite{Colson2007}. In our approach, the upper-level problem Eq. \ref{bilevel:1} determines the task segmentation and schedule, and the lower-level problem Eq. \ref{bilevel:2} determines the physical arrangement of parts in the kit.

\subsection{Task Segmentation and Scheduling}
We define our notation as follows: $K=\{a_1,a_2,...\}$ is an ordered sequence of future tasks to execute (where $a_1$ represents both robot and human tasks $a^R_1$ and $a^H_1$), $i$ is an index partitioning $K$ to determine the tasks to be included in the next kit, $T$ is a time-ordered sequence of task segments as defined above, $t \in [1,|T|]$ is the current task segment being completed, $P$ is a kit layout defined by a set of part coordinates (part id, x, y) in a kitting tray, $G$ is the task graph encoding precedence constraints for the overall task, $d$ is a scalar constant indicating the time required for a kit delivery, and $W_k$ is a scalar weight hyperparameter for the $k$th objective function component. We utilize two functions within the objective, \emph{allowed(task, set of completed tasks, task precedence graph)} which returns 1 if \emph{task} can be completed given the \emph{set of completed tasks} and the \emph{task precedence graph} (0 otherwise) and \emph{kit\_fitness(kit layout)} which evaluates a \emph{kit layout} for usability (returning a scalar where larger is better).

The optimization for timestep $t+1$ is defined as follows:
\begin{align}
\min_{K,i,P} & \text{ } W_1\sum_{j=1}^{|K|} (1-\text{allowed}(K_j, \bigcup_{x=1}^{t}T_x \cup \bigcup_{x=1}^{j-1}K_x, G)) \label{eqn: precedence constraints}\\
 & - W_2*i \label{eqn: number of tasks}\\
 & + W_3 (d + \sum_{j=1}^i \text{duration}^R(K_j) - \text{duration}^H(T_{t})) \label{eqn: idle times1}\\
 & + W_4 (d + \sum_{j=i+1}^{|K|} \text{duration}^R(K_j) - \sum_{j=1}^i \text{duration}^H(K_j)) \label{eqn: idle times2}\\
 & - W_5 * \text{kit\_fitness}(P) \label{term: lower-level}
\end{align}




The first term (\ref{eqn: precedence constraints}) allows us to heavily penalize proposed task sequences $K$ that do not satisfy precedence constraints. The second term (\ref{eqn: number of tasks}) can be used to override other terms to prioritize kits covering more tasks. The third term (\ref{eqn: idle times1}) penalizes differences between the human's remaining task time and the duration required to prepare and deliver the proposed next kit. The fourth term (\ref{eqn: idle times2}) is similar, penalizing partitions that are likely to lead to idle times after $T_{t+1}$. Finally, the fifth term (\ref{eqn: lower objective}) incentivizes kit layouts with high utility, which is solved for by the second level of optimization given a proposed $K$ and $i$.

After solving for values of $K, i, P$, we set task segment $T_{t+1} = K_{1:i}$ and increment $t$. This optimization is run on a greedy basis at each timestep while the robot executes its kitting behaviors.
%
%
Without loss of generality, $i$ can be defined as a vector of indices that create $|i|+1$ task segments $T_t = K_{0:i[0]}, \ T_{t+1} = K_{i[0]:i[1]}, \ ..., \ T_{t+|i|} = K_{i[|i|-1]:|K|}$. Terms (\ref{eqn: idle times1}, \ref{eqn: idle times2}) would be updated to minimize the differences between human and robot task times for each segment.

The proposed task segmentation and scheduling solution for JIT kitting is an instance of flow-shop scheduling, which has been shown to be NP-hard \cite{ewacha1990permutation,garey1976complexity}. We approximate the optimal solution by only considering a limited horizon and partial task plans $K$ of fixed length up to $N$ ($N=5$ in our experiments). 



\subsection{Kit Arrangement} \label{section: kit arrangement}
The nested lower-level objective function \emph{kit\_fitness} is defined to score the arrangement of parts for a given task segment on the tray used for kit delivery. This task segment $T_{t+1}$ is given by $K$ and $i$ as defined in the previous section.

Let $P$ denote the set of parts required for the tasks in $T_{t+1}$ and their coordinates on the 2D plane of the kitting tray. Let $\mathbf{P_{j,x}}$ and $\mathbf{P_{j,y}}$ denote the $x$ and $y$ coordinates of the $j$th part's centroid in the kit. These coordinates are solved for with the lower optimization, with an objective function that maximizes kit fitness.
We first define terms of the \emph{kit\_fitness} objective to prioritize logical grouping of items on the tray:

\begin{gather}
    D_{diff}(P) = \sum_{k = 1}^{|P|} \sum_{\substack{j = k + 1: \\ Q(P_j) \neq Q(P_k)}}^{|P|} \sqrt{(P_{k,x} - P_{j,x})^2 + (P_{k,y} - P_{j,y})^2} \label{eqn: D_diff}
\end{gather}
\begin{gather}
    D_{same}(P) = \sum_{k = 1}^{|P|} \sum_{\substack{j = k + 1: \\ Q(P_j) = Q(P_k)}}^{|P|} \sqrt{(P_{k,x} - P_{j,x})^2 + (P_{k,y} - P_{j,y})^2}\label{eqn: D_same}
\end{gather}where $Q(p)$ denotes the \textsl{part type} of part $p$, such as "M6 screw" or "connector." Equation \ref{eqn: D_diff} sums the Euclidean distance between all parts of different types, while \ref{eqn: D_same} does the same for all parts of the same type.

We also define a term to penalize overlap between items on a tray, as we wish to discourage the robot from piling objects on top of one another. Let $BB(p, x, y)$ denote the \textsl{bounding box} of part $p$ centered at point $(x, y)$. We define $Z$ as the total overlapping area between the bounding boxes of all parts:
\begin{equation}
	Z(P) = \sum_{k=1}^{|P|} \sum_{j=k+1}^{|P|} BB(P_k, P_{k,x}, P_{k,y}) \cap BB(P_j, P_{j,x}, P_{j,y})
\end{equation}
Lastly, we use the terms described above to define our kit fitness objective function, with terms $K,i$ defined by the upper-level optimization and $W_6$ a scalar hyperparameter to modulate the optimization's focus on discouraging part overlap:
\begin{gather}
	\min_{P} D_{same}(P) -D_{diff}(P) - W_6*Z(P)\label{eqn: lower objective} \\
    \text{subject to:} \;
    \bigcup_{k=1}^{|P|}BB(P_k, P_{k,x}, P_{k,y}) \subset \text{area}_{kit} \nonumber\\
    \bigcup_{j=1}^i \text{parts\_in}(K_j) \in P \nonumber
\end{gather}

The objective function maximizes distances of parts with different types, minimizes distances of parts with the same type, and penalizes overlaps. The constraint ensures that the 2D space taken up by parts in the kit does not exceed or extend beyond the tray area and that all necessary parts are included in the kit.

We use the cross entropy (CE) method to solve the kit arrangement problem. CE is a stochastic algorithm that provides an adaptive procedure for solving combinatorial optimization problems and has asymptotic convergence properties \cite{de2005tutorial}. Other stochastic methods such as simulated annealing \cite{van1987simulated} and genetic algorithms \cite{goldberg1988genetic} can also be used.

\begin{algorithm}
\caption{Cross Entropy Method for kit arrangement}
\label{alg:cross entropy}
\begin{algorithmic}
\Require list of parts $P$, sample count to keep $c$
\State initialize $\mu (P_{k,x}, P_{k,y}, \theta_i \; \forall \; k \in 1:|P|)$, $\Sigma$ randomly
\While{not converged and max iterations not reached}
    \State $samples \sim \mathcal{N}(\mu, \Sigma)$
    \State $scores \gets sorted(score(samples))$ 
    \Comment{score samples and sort from largest to smallest}
    
    \State $\mu \gets mean(samples[:c])$
    \State $\Sigma \gets covariance(samples[:c])$
    \Comment{recompute mean and covariance from new samples}
\EndWhile
\State \Return $\mu$
\end{algorithmic}
\end{algorithm}

Algorithm \ref{alg:cross entropy} shows the implementation for CE. We first randomly initialize the mean vector and covariance matrix. The mean vector contains the position ($x$ and $y$) and orientation ($\theta$) of the parts that have to be placed on the tray. While the solution hasn't converged or the maximum number of iterations (100) has not been reached, we draw samples from a Gaussian distribution ($200$ in our experiments) with the current estimated mean and covariance. The samples are scored using Equation \ref{eqn: lower objective} and sorted from largest to smallest. Then we use the top $c$ samples to recompute the mean and covariance ($c=30$ in our experiments). The algorithm returns the estimated mean after convergence or after the maximum number of iterations.


\begin{figure*}
\centering
    \begin{subfigure}[h]{0.24\textwidth}
         \centering
         \includegraphics[width=\textwidth]{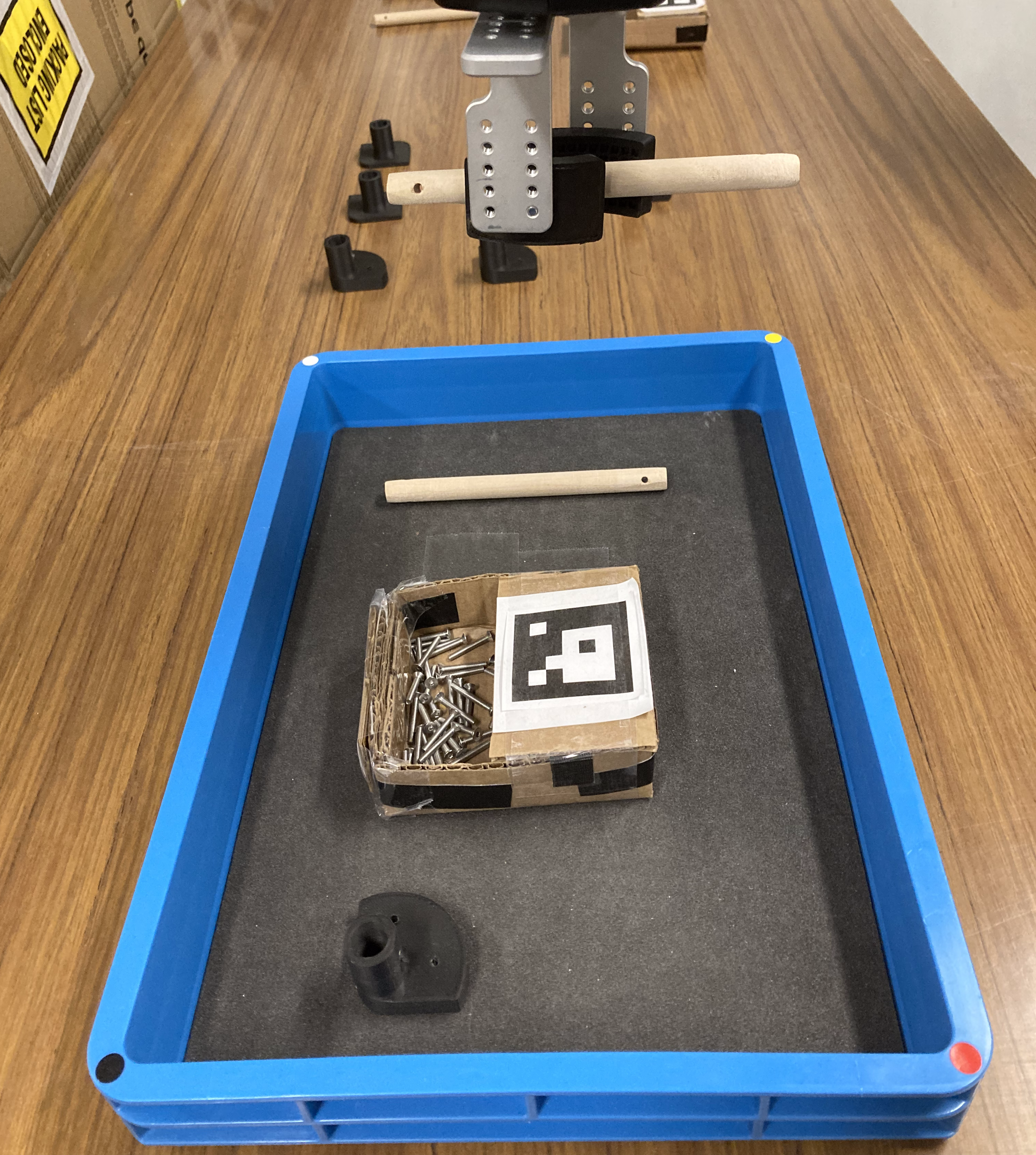}
         \caption{}
         \label{fig: arranging}
    \end{subfigure}
    \begin{subfigure}[h]{0.24\textwidth}
         \centering
         \includegraphics[width=\textwidth]{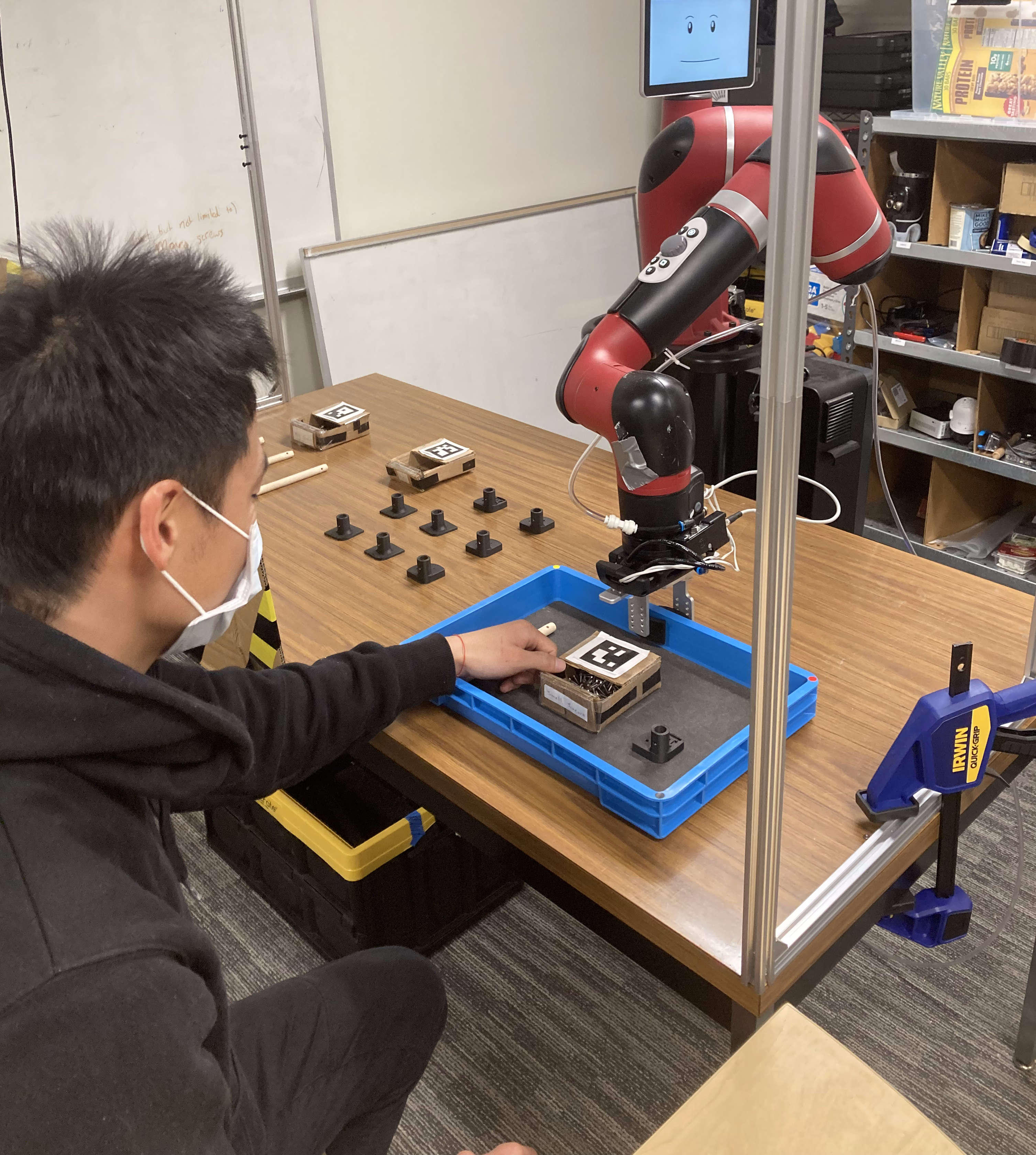}
         \caption{}
         \label{fig: delivery}
     \end{subfigure}
     \begin{subfigure}[h]{0.24\textwidth}
         \centering
         \includegraphics[width=\textwidth]{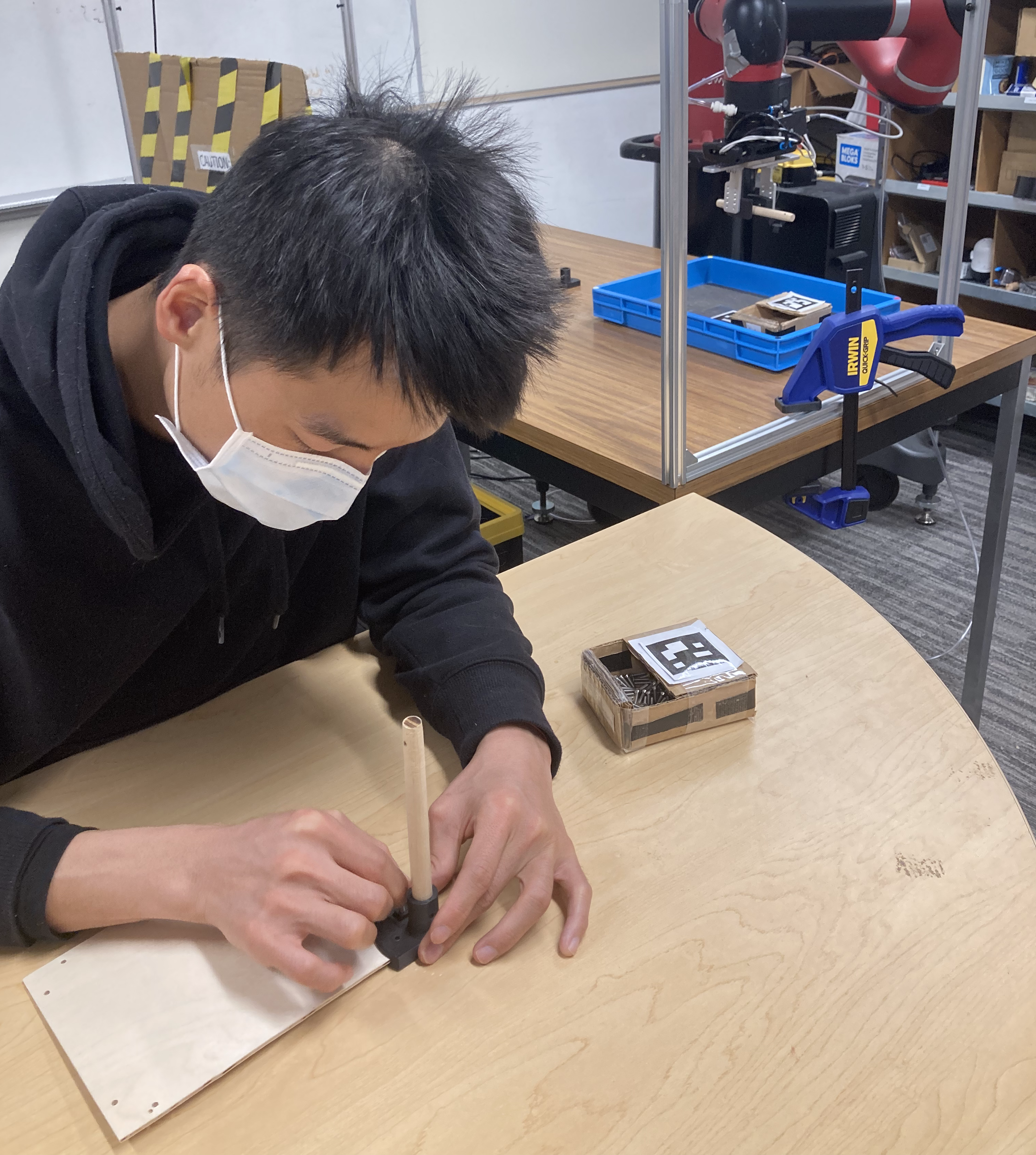}
         \caption{}
         \label{fig: building}
     \end{subfigure}
    \begin{subfigure}[h]{0.24\textwidth}
         \centering
         \includegraphics[width=\textwidth]{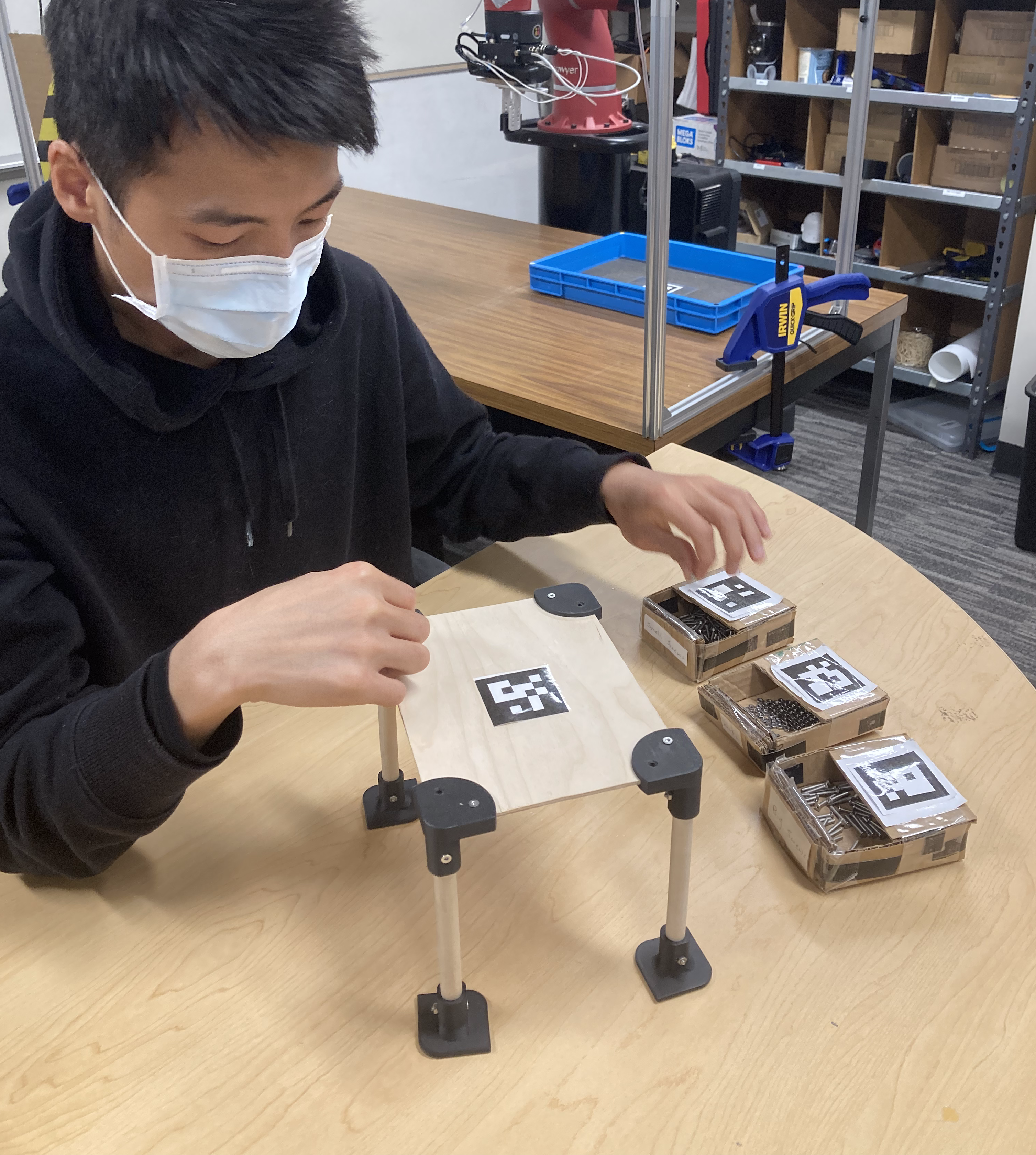}
         \caption{}
         \label{fig: finished}
     \end{subfigure}
\caption{(a) The robot arranges parts on the tray according to the lower-level optimization. (b) The robot delivers a finished kit to the user. (c) The user begins the assembly task while the robot prepares the kit for the next tasks. (d) The user finishes assembling a flat-pack furniture table.}
\label{fig: table}
\end{figure*}


\section{Experimental Validation}\label{section: experimental validation}

With the goal of evaluating our proposed system under a variety of logistic delay scenarios, we first perform a user study to collect user experience data. We then use this study as a basis to perform a series of simulated shop floor evaluations of the robot's kitting delivery performance under multiple conditions varying supply failure likelihood and delivery distance between kitting and assembly stations.

\subsection{User Study} \label{section: user study}
We had 19 people (10 males, 9 females) participate in our IRB-approved study, ranging in age from 22 to 31 ($M=25.47$, $SD=2.37$). One data point was discarded because the Sawyer robot lost connection to the study computer in the middle of the experiment. Participants were tasked with the construction of a flat-pack furniture table \cite{zeylikman2018hrc} shown in Figure \ref{fig: table}. Table assembly involves connecting four feet, four legs, four connectors, and a flat surface plank by snapping the pieces together and securing with screws and nuts.

\subsubsection{Experimental Design}
We use a within-subjects experiment to evaluate our optimization approach and two baseline conditions: (a) \textsl{Whole Assembly} is a baseline which resembles traditional kitting strategies where all the required parts for a single unit are placed on the kitting tray prior to assembly, (b) \textsl{Single Task} is a human-designed JIT schedule where the robot delivers parts for a single task at a time as segmented by the task graph itself (one kit per vertex), and (c) \textsl{Optimized} is the result of applying the proposed bilevel optimization algorithm to the assembly task. To minimize learning effects, we counterbalance the condition orderings and ask the participants to practice assembling part of the table before the experiment.

The assembly task is divided into a total of $12$ tasks, i.e. $3$ tasks per leg of the table. The precedence constraints (see Section \ref{section: problem formulation}) dictate that the feet and connecting joint have to be secured to the leg before the leg can be secured to the plank. The robot's tasks include delivering the screws, nuts, four legs, four feet, and four joints to the human participant via a kitting tray. The human participant and the robot have separate workspaces, and the human participant is instructed to not enter the robot's workspace during the experiment. Figure \ref{fig: experiment setup} shows the experiment setup.

\subsubsection{Metrics}
The objective metrics used to assess the efficiency of our proposed approach are the total task time and the human idle time. The \textsl{total task time} is the total experiment time for assembling a single table. The \textsl{human idle time} is defined as the duration of the experiment in which the participant was not actively assembling parts.
A post-experiment survey was used to assess the user experience: participants were asked to rank each trial based on the robot's Usefulness, Intuitiveness, and Efficiency during that trial.

\subsubsection{Hypotheses}

\begin{itemize}
    \item \textbf{H1:} The \textsl{Optimized} condition will have a makespan lower-bounding \textsl{Single Task} and \textsl{Whole Assembly} conditions with reduced idle time that is proportional to delivery duration.
    \item \textbf{H2:} Participants will perceive the \textsl{Optimized} condition more highly along subjective measures (usefulness, intuitiveness, and efficiency) than the \textsl{Single Task} and \textsl{Whole Assembly} conditions.
\end{itemize}

\begin{figure}
    \begin{subfigure}[ht]{0.85\linewidth}
        \includegraphics[width=\textwidth]{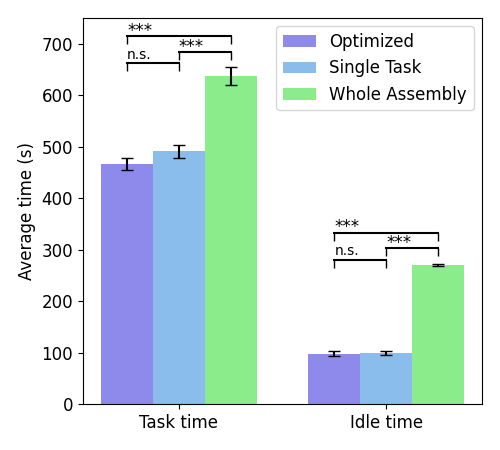}
    \end{subfigure} \vspace{-3pt}
    \begin{subfigure}[ht]{0.85\linewidth}
        \includegraphics[width=\textwidth]{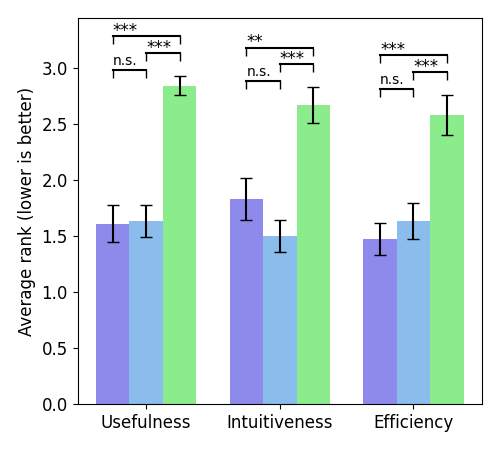}
    \end{subfigure}
    \centering
    
    \caption{\textbf{\textsl{Top:}} Total task time and human idle time across all three conditions. Both the \textsl{Optimized} and \textsl{Single Task} approaches had significantly lower total times in both metrics than in the \textsl{Whole Assembly} condition. \textbf{\textsl{Bottom:}} Total user ranking scores for each condition across three subjective metrics: usefulness, intuitiveness, and efficiency. The \textsl{Whole Assembly} condition had significantly worse scores than the algorithmic conditions across all three rankings. The \textsl{Single Task} and \textsl{Optimized} conditions did not vary significantly on any of the above metrics, likely due to the proximity of the robot to the human (no delivery delays).\\ ** indicates $p < .001$; *** indicates $p < .0001$. }
    \label{fig: userstudy}
\end{figure}

\subsubsection{Study Protocol}
After a brief introduction to the robot's role in part delivery, the human participant was presented with a printed guide document, which included step-by-step instructions in both written and visual form. The participant was asked to practice constructing a single leg of the table before the experiment began. 

During the experiment, the required parts were initially placed in the robot workspace. The robot kitted the assembly parts by placing some number of them onto a tray, and then pushing the tray towards the human. The participant was then notified to retrieve the parts on the tray. The robot returned to its kitting task once the participant retrieved the parts. The same assembly was repeated for all three conditions. 

\subsubsection{Implementation Details}
We implemented the optimization framework on a Sawyer Research Robot, using the Robot Operating System (ROS) \cite{koubaa2017robot}. We installed an Intel RealSense D435 Depth Camera above the robot workspace and used the Facebook Detectron2 algorithm \cite{wu2019detectron2} for image segmentation of the assembly parts. We used the ArUco library \cite{aruco2} to detect markers placed on the screw boxes and tray to read their pose.

\subsection{Shop Floor Simulation}

While the user study provides useful insights for user preferences and task efficiency for a single assembly, we use discrete event simulation (DES) to evaluate our proposed approach when multiple assemblies of the furniture table are performed in sequence. We evaluate the framework under various assembly part arrival time distributions and part-feeding machine breakdown conditions (i.e. the part is not available to the robot until after the machine is repaired). DES is able to explicitly model events using probability distributions and answers questions such as the line throughput and utilization \cite{prajapat2017review}.

\subsubsection{Experimental Design}

We simulated the assembly of ten flat-pack furniture tables. For each assembly, we randomly sampled a participant's task times and the robot's task times as recorded from the user study detailed in \ref{section: user study}. The arrival time of assembly parts to the robot's workspace is modeled by an exponential distribution with Mean Arrival Time denoted by $MAT$. In other words, the arrival of parts is a Poisson process with rate $1/MAT$. Since increasing the arrival times of all the parts linearly increases the total task times, we only vary the $MAT$ of the leg and foot part types. We also model the machine breakdown for the feeding of the leg and foot parts with an exponential distribution. The mean time to failure of the machines is denoted by $MTTF$. Each machine has a fixed repair time of $30$ seconds, after which it starts feeding parts again according to the part arrival process with rate $1/MAT$.


\subsubsection{Hypothesis}
\begin{itemize}
    \item \textbf{H3:} When simulating multiple assemblies of the task, the \textsl{Optimized} condition will significantly outperform baseline kitting strategies in both total task time and human idle time in the presence of logistical delays.
\end{itemize}

\subsubsection{Implementation Details}
The simulation is implemented in Python using the SimPy library. In order to account for assembly part shortages due to slow arrival times or machine breakdowns, we add an additional term to the upper-level objective function:
\begin{equation}
    W_7 \sum_{k=1}^{|P|} U(P_k)
\end{equation}

where $U(p)$ returns 1 if part $p$ is currently unavailable for the robot to place on the tray and 0 otherwise. 

\begin{figure*}
    \centering
    \begin{subfigure}[ht]{0.47\textwidth}
        \includegraphics[width=\textwidth]{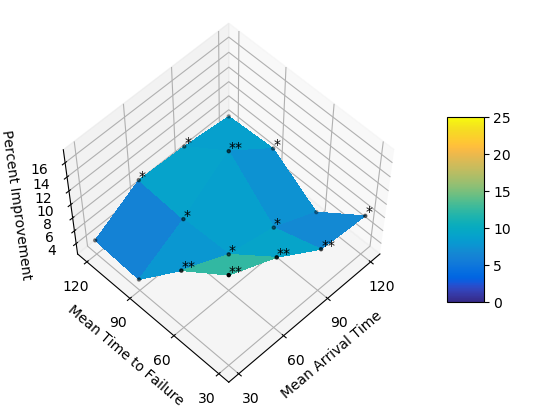}
        \caption{Percent improvement in total task time of \textsl{Optimized} over \textsl{Single Task}.}
        \label{fig: tt_optimized_over_single}
    \end{subfigure}
    \hfill
    \begin{subfigure}[ht]{0.47\textwidth}
        \includegraphics[width=\textwidth]{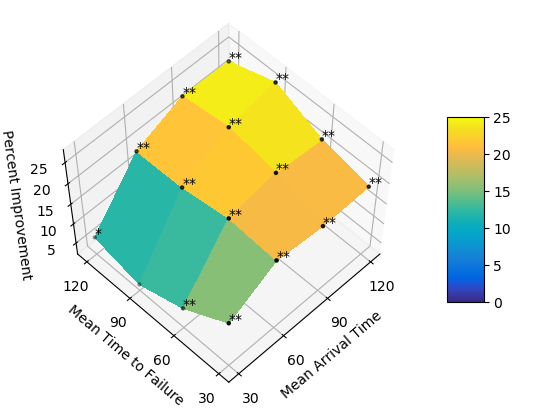}
        \caption{Percent improvement in total task time of \textsl{Optimized} over \textsl{Whole Assembly}.}
        \label{fig: tt_optimized_over_whole}
    \end{subfigure}
    
    \begin{subfigure}[ht]{0.47\textwidth}
        \includegraphics[width=\textwidth]{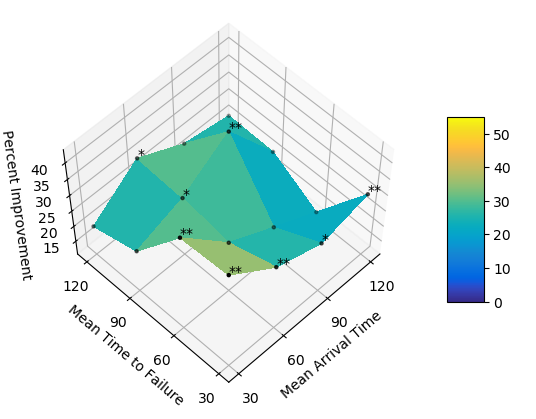}
        \caption{Percent improvement in human idle time of \textsl{Optimized} over \textsl{Single Task}.}
        \label{fig: hit_optimized_over_single}
    \end{subfigure}
    \hfill
    \begin{subfigure}[ht]{0.47\textwidth}
        \includegraphics[width=\textwidth]{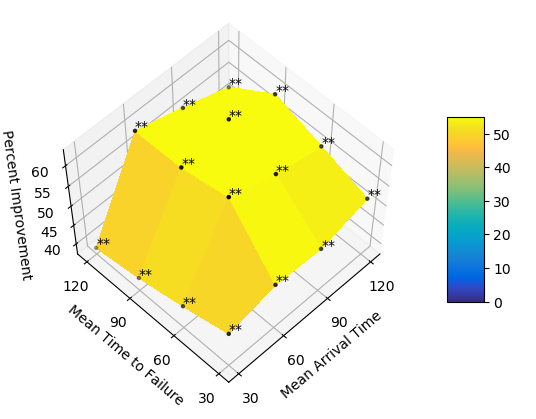}
        \caption{Percent improvement in human idle time of \textsl{Optimized} over \textsl{Whole Assembly}.}
        \label{fig: hit_optimized_over_whole}
    \end{subfigure}\vspace{-4pt}
    \caption{Comparisons of Optimized approach with baselines in simulation. (a) and (b) show the percent improvement in total task time of the \textsl{Optimized} approach over the \textsl{Single Task} and \textsl{Whole Assembly} conditions respectively. With low part shortage (i.e. low \textsl{MAT}), \textsl{Optimized} is most advantageous over \textsl{Single Task} because \textsl{Single Task} has many more tray deliveries which increase the total robot task time. With high part shortage (i.e. high \textsl{MAT}), \textsl{Optimized} is most advantageous over \textsl{Whole Assembly} because \textsl{Optimized} adapts the kitting strategy according to which parts are available. (c) and (d) show the percent improvement in human idle time of the \textsl{Optimized} approach over the \textsl{Single Task} and \textsl{Whole Assembly} conditions respectively. The trend for human idle time is similar to that of total task time. * denotes $p < .05$ while ** denotes $p < .001.$}
\end{figure*}

\section{Results}\label{section: results}

\subsection{User Study}\label{sec:userstudy}
\subsubsection{Objective measures}
Total task time and human idle time measures were analyzed using the one-way analysis of variances (ANOVA) with experimental condition as an independent variable. Post-hoc analysis used Tukey's HSD test for multiple comparisons to test for effects between condition pairs. The results are summarized in Figure \ref{fig: userstudy}.

\paragraph*{Total task time} The effect of condition significantly influenced total task time, $F(2, 34) = 57.60, p < .0001$, with a significant reduction in task time between the \textsl{Whole Assembly} condition and the \textsl{Single Task} condition ($p < .0001$, 95\% C.I. = [-188.211, -103.851]), as well as between the \textsl{Whole Assembly} and \textsl{Optimized} conditions ($p < .0001$, 95\% C.I. = [-213.213, -128.853]).

\paragraph*{Human idle time} The effect of experimental condition significantly influenced human idle time, $F(2, 34) = 709.87, p < .0001$; there was a significant reduction in idle time between the \textsl{Whole Assembly} and \textsl{Single Task} conditions ($p < .0001$, 95\% C.I. = [-184.465, -158.586]), as well as between the \textsl{Whole Assembly} and \textsl{Optimized} conditions ($p < .0001$, 95\% C.I. = [-186.020, -160.141]).

\subsubsection{Subjective measures}
 We analyzed rank scale data on our post-experiment survey using a nonparametric Kruskal-Wallis Test with experimental condition as a fixed effect. Post-hoc comparisons used the Wilcoxon method for analyzing rank significance between condition pairs. The results are summarized in Figure \ref{fig: userstudy}. 

\subsection{Simulation}
In simulation, we compare the task time and human idle time outcomes between \textsl{fixed} strategies (\textsl{Single Task} and \textsl{Whole Assembly}) and the \textsl{Optimized} strategy. Percent improvements in task time and human idle time between conditions are summarized in Figure 6. Results were analyzed using the same methods as in Section \ref{sec:userstudy}.



\section{Discussion}\label{section: discussion}

Hypothesis \textbf{H1} regarding task time and idle time is supported by the user study results. We suspect that \textsl{Optimized} is similar to \textsl{Single Task} due to the high variance in the participants' task times and the absence of delivery distance (the human and robot were co-located across a table). The average human task time (i.e. non-idle time) for a single assembly and across all conditions was $374.78$ seconds with a standard deviation of $61.14$ seconds. 

Hypothesis \textbf{H2} is supported; the \textsl{Optimized} condition had significantly lower ranking scores (i.e. were ranked better) in usefulness, intuitiveness, and efficiency than \textsl{Whole Assembly}. However, there was no significant difference in rankings of \textsl{Single Task} and \textsl{Optimized} on any of the three subjective metrics (all $p \geq .05$). From the post-experiment interviews, participants found the \textsl{Optimized} or \textsl{Single Task} conditions most efficient because \textsl{"it gave me the parts in timely manner"}, and \textsl{"[it has] less waiting time."} This suggests that participants preferred just-in-time kitting strategies.
%
%
Some participants indicated they found the just-in-time approaches less cognitively demanding; \textsl{"[The robot is] guiding me through the most efficient way to complete the assembly job ... the robot lifted part of my pressure to plan it."}

Hypothesis \textbf{H3} is supported; the total task time and human idle time are significantly shorter for \textsl{Optimized} than either \textsl{Single Task} or \textsl{Whole Assembly} for most of the scenarios with logistic delays. \textsl{Optimized} is most advantageous over \textsl{Whole Assembly} when there is a high part shortage (high \textsl{MAT}) because \textsl{Whole Assembly} waits for all the parts to be kitted before delivering the kit to the assembly station. When a part is delayed, \textsl{Whole Assembly} waits until the part becomes available. While \textsl{Single Task} also suffers from logistic delays, the effect is less prominent because kits are delivered with partial assembly parts. The \textsl{Optimized} approach is able to dynamically determine which tasks to deliver parts for in order to minimize idle times and maximize efficiency.

\section{Conclusions}\label{section: conclusions}

This work introduces a bilevel optimization approach for robot kitting and demonstrates its ability to reduce both overall task time and human and robot idle times for a furniture assembly task. In a user study, we evaluated this approach against a generic whole-kit assembly (\textsl{Whole Assembly}) and a human-designed just in time approach (\textsl{Single Task}), and found that just in time kitting had quicker task completion and were rated more highly by users on a number of subjective metrics. Simulating longer and more varied task environments revealed that the online optimized approach demonstrates significant performance improvement over the fixed kitting strategies in more realistic settings that include logistic delays.

The user study suggested two additional avenues of future research: online estimates of human task time to better minimize idle times and the addition of human factors into the optimization framework. Lastly, we plan to investigate which factors best improve the efficiency and intuitiveness of kitting tray designs. One limitation of this work is the limited horizon used which may not generate the most optimal kitting strategy with respect to the objective function. With more compute power or longer planning times, our approach can generate more optimal kits.

\bibliographystyle{IEEEtranS}
\bibliography{references}

\begin{thebibliography}{10}
\providecommand{\url}[1]{#1}
\csname url@samestyle\endcsname
\providecommand{\newblock}{\relax}
\providecommand{\bibinfo}[2]{#2}
\providecommand{\BIBentrySTDinterwordspacing}{\spaceskip=0pt\relax}
\providecommand{\BIBentryALTinterwordstretchfactor}{4}
\providecommand{\BIBentryALTinterwordspacing}{\spaceskip=\fontdimen2\font plus
\BIBentryALTinterwordstretchfactor\fontdimen3\font minus
  \fontdimen4\font\relax}
\providecommand{\BIBforeignlanguage}[2]{{%
\expandafter\ifx\csname l@#1\endcsname\relax
\typeout{** WARNING: IEEEtranS.bst: No hyphenation pattern has been}%
\typeout{** loaded for the language `#1'. Using the pattern for}%
\typeout{** the default language instead.}%
\else
\language=\csname l@#1\endcsname
\fi
#2}}
\providecommand{\BIBdecl}{\relax}
\BIBdecl

\bibitem{boudella2018kitting}
M.~E.~A. Boudella, E.~Sahin, and Y.~Dallery, ``Kitting optimisation in
  just-in-time mixed-model assembly lines: assigning parts to pickers in a
  hybrid robot--operator kitting system,'' \emph{International Journal of
  Production Research}, vol.~56, no.~16, pp. 5475--5494, 2018.

\bibitem{bozer1992kitting}
Y.~A. Bozer and L.~F. McGinnis, ``Kitting versus line stocking: A conceptual
  framework and a descriptive model,'' \emph{International Journal of
  Production Economics}, vol.~28, no.~1, pp. 1--19, 1992.

\bibitem{caputo2021model}
A.~C. Caputo, P.~M. Pelagagge, and P.~Salini, ``A model for planning and
  economic comparison of manual and automated kitting systems,''
  \emph{International Journal of Production Research}, vol.~59, no.~3, pp.
  885--908, 2021.

\bibitem{Colson2007}
\BIBentryALTinterwordspacing
B.~Colson, P.~Marcotte, and G.~Savard, ``An overview of bilevel optimization,''
  \emph{Annals of Operations Research}, vol. 153, no.~1, pp. 235--256, Sep
  2007. [Online]. Available: \url{https://doi.org/10.1007/s10479-007-0176-2}
\BIBentrySTDinterwordspacing

\bibitem{de2005tutorial}
P.-T. De~Boer, D.~P. Kroese, S.~Mannor, and R.~Y. Rubinstein, ``A tutorial on
  the cross-entropy method,'' \emph{Annals of operations research}, vol. 134,
  no.~1, pp. 19--67, 2005.

\bibitem{ewacha1990permutation}
K.~Ewacha, I.~Rival, and G.~Steiner, ``Permutation schedules for flow shops
  with precedence constraints,'' \emph{Operations research}, vol.~38, no.~6,
  pp. 1135--1139, 1990.

\bibitem{fansuri2017challenges}
A.~Fansuri, A.~Rose, N.~N. Mohamed, and H.~Ahmad, ``The challenges of lean
  manufacturing implementation in kitting assembly,'' in \emph{IOP Conference
  Series: Materials Science and Engineering}, vol. 257, no.~1.\hskip 1em plus
  0.5em minus 0.4em\relax IOP Publishing, 2017, p. 012069.

\bibitem{garey1976complexity}
M.~R. Garey, D.~S. Johnson, and R.~Sethi, ``The complexity of flowshop and
  jobshop scheduling,'' \emph{Mathematics of operations research}, vol.~1,
  no.~2, pp. 117--129, 1976.

\bibitem{aruco2}
S.~Garrido-Jurado, R.~Muñoz-Salinas, F.~Madrid-Cuevas, and R.~Medina-Carnicer,
  ``Generation of fiducial marker dictionaries using mixed integer linear
  programming,'' \emph{Pattern Recognition}, vol.~51, 10 2015.

\bibitem{goldberg1988genetic}
D.~E. Goldberg and J.~H. Holland, ``Genetic algorithms and machine learning,''
  \emph{Machine Learning}, 1988.

\bibitem{jeanson2018tell}
L.~Jeanson, J.~C. Bastien, A.~Morais, and J.~Barcenilla, ``Tell me how you kit,
  i’ll tell you how you think,'' in \emph{H-Workload 2018: 2nd International
  Symposium on Human Mental Workload: Models and Applications}, 2018, p. 232.

\bibitem{joundi2019understanding}
J.~Joundi, P.~Conradie, J.~Van Den~Bergh, and J.~Saldien, ``Understanding and
  exploring operator needs in mixed model assembly,'' in \emph{EISMS19,
  Workshop on Research and Practice Challenges for Engineering Interactive
  Systems while Integrating Multiple Stakeholders Viewpoints}, 2019.

\bibitem{koubaa2017robot}
A.~Koubaa, \emph{Robot Operating System (ROS): The Complete Reference (Volume
  2)}, 1st~ed.\hskip 1em plus 0.5em minus 0.4em\relax Springer Publishing
  Company, Incorporated, 2017.

\bibitem{maderna2020online}
R.~Maderna, M.~Poggiali, A.~M. Zanchettin, and P.~Rocco, ``An online scheduling
  algorithm for human-robot collaborative kitting,'' in \emph{2020 IEEE
  international conference on robotics and automation (ICRA)}.\hskip 1em plus
  0.5em minus 0.4em\relax IEEE, 2020, pp. 11\,430--11\,435.

\bibitem{time-aware-multiagent}
\BIBentryALTinterwordspacing
M.~Maniadakis, E.~Hourdakis, M.~Sigalas, S.~Piperakis, M.~Koskinopoulou, and
  P.~Trahanias, ``Time-aware multi-agent symbiosis,'' \emph{Frontiers in
  Robotics and AI}, vol.~7, 2020. [Online]. Available:
  \url{https://www.frontiersin.org/article/10.3389/frobt.2020.503452}
\BIBentrySTDinterwordspacing

\bibitem{nikam2018design}
M.~S. Nikam, A.~Joel, and J.~Dutta, ``Design and implementation of kitting
  trolley for just in time production in textile industry,''
  \emph{International Research Journal of Engineering and Technology (IRJET)},
  vol.~5, no.~4, pp. 1301--1304, 2018.

\bibitem{multi-robot-auction}
E.~Nunes and M.~Gini, ``Multi-robot auctions for allocation of tasks with
  temporal constraints,'' in \emph{Proceedings of the Twenty-Ninth AAAI
  Conference on Artificial Intelligence}, ser. AAAI'15.\hskip 1em plus 0.5em
  minus 0.4em\relax AAAI Press, 2015, p. 2110–2216.

\bibitem{prajapat2017review}
N.~Prajapat and A.~Tiwari, ``A review of assembly optimisation applications
  using discrete event simulation,'' \emph{International Journal of Computer
  Integrated Manufacturing}, vol.~30, no. 2-3, pp. 215--228, 2017.

\bibitem{shaikh2012effects}
S.~Shaikh, S.~Cobb, D.~Golightly, J.~Segal, and C.~Haslegrave, ``Effects of
  takt time on physical and cognitive demands in a mixed model assembly line
  and a single model assembly line,'' in \emph{Contemporary Ergonomics and
  Human Factors 2012: Proceedings of the international conference on Ergonomics
  \& Human Factors 2012, Blackpool, UK, 16-19 April 2012}.\hskip 1em plus 0.5em
  minus 0.4em\relax CRC Press, 2012, p. 137.

\bibitem{townsend2012line}
B.~Townsend, \emph{Line Balancing And Jit Kitting}.\hskip 1em plus 0.5em minus
  0.4em\relax Boca Raton, Fl: Crc Press, 2012.

\bibitem{van1987simulated}
P.~J.~M. van Laarhoven and E.~H.~L. Aarts, ``Simulated annealing: Theory and
  applications,'' in \emph{Mathematics and Its Applications}, 1987.

\bibitem{wilcox2013optimization}
R.~Wilcox, S.~Nikolaidis, and J.~Shah, ``Optimization of temporal dynamics for
  adaptive human-robot interaction in assembly manufacturing,''
  \emph{Robotics}, vol.~8, no. 441, pp. 10--15\,607, 2013.

\bibitem{wu2019detectron2}
Y.~Wu, A.~Kirillov, F.~Massa, W.-Y. Lo, and R.~Girshick, ``Detectron2,''
  \url{https://github.com/facebookresearch/detectron2}, 2019.

\bibitem{zeylikman2018hrc}
S.~Zeylikman, S.~Widder, A.~Roncone, O.~Mangin, and B.~Scassellati, ``The hrc
  model set for human-robot collaboration research,'' in \emph{2018 IEEE/RSJ
  International Conference on Intelligent Robots and Systems (IROS)}.\hskip 1em
  plus 0.5em minus 0.4em\relax IEEE, 2018, pp. 1845--1852.

\bibitem{zhou2021static}
B.~Zhou and Z.~He, ``A static semi-kitting strategy system of jit material
  distribution scheduling for mixed-flow assembly lines,'' \emph{Expert Systems
  with Applications}, vol. 184, p. 115523, 2021.

\bibitem{zhu2008}
\BIBentryALTinterwordspacing
X.~Zhu, S.~J. Hu, Y.~Koren, and S.~P. Marin, ``{Modeling of Manufacturing
  Complexity in Mixed-Model Assembly Lines},'' \emph{Journal of Manufacturing
  Science and Engineering}, vol. 130, no.~5, 08 2008, 051013. [Online].
  Available: \url{https://doi.org/10.1115/1.2953076}
\BIBentrySTDinterwordspacing

\end{thebibliography}

\end{document}